\documentclass[fleqn]{article}
\usepackage{ijcai15}

\usepackage{graphicx}

\usepackage{times}
\usepackage{amsmath, amsthm}
\usepackage{txfonts}

\newtheorem{lemma}{Lemma}
\newtheorem{theorem}{Theorem}

\newcommand{\lang}{$\mathcal{ALCH+}$ }
\newcommand{\compbit}{$b_\omega$}
\newcommand{\bsim}{$\mathcal{\sigma_{BS}}$ }
\newcommand{\bcapsim}{$\hat{\sigma}_{\mathcal{BS}}$}
\newcommand{\isim}{$\mathcal{I_B}$ }
\newcommand{\interpret}{$^\mathcal{I}$ }
\newcommand{\lsim}{$\mathcal{L_B}$ }
\newcommand{\alphsim}{$\Sigma_\mathcal{B}$}
\newcommand{\bunion}{$\sqcup_\mathcal{B}$ }
\newcommand{\bintersect}{$\sqcap_\mathcal{B}$ }
\newcommand{\bcode}{$\omega_\mathcal{B}$ }
\newcommand{\bcodebase}{$\omega_{\mathcal{B}_{base}}$}
\newcommand{\bcodederived}{$\omega_{\mathcal{B}_{derived}}$}
\newcommand{\bconcat}{\circ_\mathcal{B}}
\newcommand{\concat}{\circ}
\newcommand{\border}{$\preccurlyeq_\mathcal{B}$}
\newcommand{\bsimorder}{$\preccurlyeq_{\sigma_\mathcal{BS}}$}
\newcommand{\regbcodederived}{(\concat\langle(\omega_{\mathcal{B}_{derived}})\concat\rangle)^\ast}

\title{\textit{BitSim}: An Algebraic Similarity Measure for Description Logics Concepts}
\author{Description Logics, Semantic Similarity}

\begin{document}

\maketitle

\begin{abstract}
  In this paper, we propose an algebraic similarity measure $\mathcal{\sigma_{BS}}$ ($\mathcal{BS}$ stands for \textit{BitSim}) for assigning semantic similarity score to concept definitions in $\mathcal{ALCH+}$ - an expressive fragment of Description Logics (DL). We define an algebraic interpretation function, $\mathcal{I_{B}}$, that maps a concept-definition to a unique string ($\omega_\mathcal{B}$) (called \textit{bit-code}) over an alphabet $\mathcal{\sum_{B}}$ of 11 symbols belonging to $\mathcal{L_{B}}$ - the language over $\mathcal{\sum_{B}}$. $\mathcal{I_B}$ has semantic correspondence with conventional model-theoretic interpretation of DL. We then define $\mathcal{\sigma_{BS}}$ on $\mathcal{L_{B}}$. A detailed analysis of $\mathcal{I_B}$ and $\mathcal{\sigma_{BS}}$ has been given.
\end{abstract}

\section{Introduction}

Semantic similarity measure serves as the foundation of knowledge discovery and management processes such as ontology matching, ontology alignment \& mapping, ontology merging, etc \cite{shvaiko2013ontology}. Ontological concept similarity can be based on different approaches: (i) string matching of concept labels (i.e. \textit{lexical similarity}) \cite{stoilos2005string}, (ii) external lexical resource/ontology based matching (i.e. \textit{lexico-semantic similarity}) \cite{rada1989development}, (iii)  graph-based matching using lexicons such as WordNet \cite{stuckenschmidt2007partial} (i.e. \textit{structural similarity}), (iv) property analysis (as in FCA-based similarity \cite{cimiano2005learning}) or instance analysis (as in Jaccard similarity \cite{jaccard}) based matching over a large sample of concept instance occurrences  (i.e. \textit{instance-driven similarity}), (v) matching based on statistical analysis of attribute-value or distribution analysis within fixed context-windows of concepts over large corpora (i.e. \textit{statistical similarity}) \cite{li1994semantic}, and (vi) model-theoretic matching of formal concept descriptions (i.e. \textit{formal semantic similarity}) \cite{alsubait2014measuring}. 

It can be argued that, in comparison to other approaches, formal semantic similarity measure modeling has not received equal research attention. Nevertheless, existing literature is significant, and can be broadly classified into two approaches: (i) Propositional Logics based \cite{nienhuys1998distances,ramon1998framework}, and (ii) Description Logics (DL) based \cite{alsubait2014measuring,lehmann2012framework,stuckenschmidt2007partial,fanizzi2006similarity,borgida2005towards}. The former requires: (a) representation of ontologies (mostly in RDFS/OWL format) in First Order Predicate Logic, (b) a set of \textit{axioms} (or domain knowledge, mostly as upper ontologies/thesaurus), and (c) a SAT solver that checks satisfiability (and hence, satisfiability) of disjointness of concept pairs. The latter approach, on the other hand, does not necessarily require any formal language transformation or satisfiability checker. In this paper we propose an algebraic similarity measure, called \textit{BitSim} ($\sigma_{\mathcal{BS}}$), that can compute semantic similarity of pair of concepts defined in $\mathcal{ALCH+}$\footnote{\lang: $\mathcal{ALCH} \cup \{\mathcal{R}_\mathcal{UNION}, \mathcal{R}_\mathcal{INTERSECTION}\}$}. The motivation behind $\sigma_{\mathcal{BS}}$ is to formulate a formal semantic similarity measure that provides: (i) a platform for fast, scalable, and accurate semantic similarity computation of DL concepts, and (ii) a  sound and complete correspondence with conventional semantic interpretation of DL. $\sigma_{\mathcal{BS}}$ is algebraic, in the sense that it maps a given pair of concept codes (called \textit{bit-code}), instead of concept DL definitions/axioms, to a positive real space. For this we define a novel algebraic interpretation function, called $\mathcal{I_B}$, that maps an $\mathcal{ALCH+}$ definition to a unique string, called \textit{bit-code}, ($\omega_\mathcal{B}$) belonging to the language $\mathcal{L_B}$ defined over a novel algebraic alphabet $\sum_{\mathcal{B}}$. We prove that \isim has complete correspondence with $\mathcal{I}_{\mathcal{ALCH+}}$. We also show that \bsim is highly adaptive to any kind of similarity measure that relies on set operation. As an example, we have shown how \bsim can be plugged into Jaccard similarity index. The contribution of the paper is as follows:
\begin{itemize}
	\item \isim: A novel algebraic semantic interpretation function for \lang.
	\item Proof of mathematical correspondence of \isim with semantic interpretation of \lang.
	\item \bsim: A novel semantic similarity measure based on \isim
	\item Comparative analysis of properties of \bsim with contemporary DL based similarity measures.	
\end{itemize}

\section{Related Work}
DL based similarity measures, as described in the introduction, can be further sub-divided into: (i) taxonomic analysis, (ii) structural analysis \cite{tongphudesirable,ontanon2012similarity,Joslyn2008,Hariri2006}, (iii) language approximation \cite{stuckenschmidt2007partial,tserendorj2008approximate,groot2005approximating,BrandtKT02}, and (iii) model-theoretic analysis \cite{distel2014concept,alsubait2014measuring,lehmann2012framework,borgida2005towards}. The most common approach for DL based similarity measure modeling adopts taxonomic analysis as proposed in \cite{Rada1989,resnik1999semantic,jiang1997semantic,wu1994verbs,lin1998information}. These techniques can be further sub-divided, as mentioned in introduction, into graph-traversal approaches \cite{Rada1989} and Information-Content approaches \cite{resnik1999semantic}. However, these methods can work on a generalized ontology and hence, are not sensitive to DL definitions. 

In structural analysis based approaches, a similarity measure is designed to capture the semantic equivalence of description trees of definitions of DL concept pairs. One way of achieving this is to calculate the degree of homomorphism between such trees, as proposed in \cite{tongphudesirable}. A refinement graph based anti-unification approach has been proposed in \cite{ontanon2012similarity} for computing instance similarity. Approximation based techniques  aim at converting given DL expression to another lower expressive DL language on approximation. In \cite{stuckenschmidt2007partial} an upper and lower approximation interpretation for $\mathcal{SHIQ}$ have been defined over a sub-vocabulary of $\mathcal{SHIQ}$. The sub-vocabulary can be formed by either removing concepts atoms in a given definition or by replacing them with structurally simpler concepts \cite{groot2005approximating}. Another technique, as proposed in \cite{Noia2004}, is based on converting user query into a DL expression and try to classify the match to be either an \textit{exact match}, or a \textit{potential match} (i.e., match might happen if some concept atom and operators are added) or a \textit{partial match} (where the user query and answer/description found in the knowledge base are in conflict). 

One of the pioneer work on model-theoretic interpretation based similarity approach can be found in \cite{borgida2005towards}. The work shows the inherent difficulty in measuring similarity of DL concepts using conventional taxonomic analysis based techniques. It then uses an Information-Content based approach to evaluate the similarity of two concept definitions. A work has been proposed by \cite{lehmann2012framework} for similarity computation of concepts defined in $\mathcal{ELH}$. In this work, a Jaccard Index \cite{jaccard} based approach has been followed that compares common parents of a concept pair using a \textit{fuzzy connector} (i.e. a similarity score aggregation function). A similar Jaccard Index based approach has been recently proposed in \cite{alsubait2014measuring}. Another recent approach has been proposed in \cite{distel2014concept}. The work emphasizes the necessity of \textit{triangle inequality} property of formal semantic similarity measure. It defines two versions of a \textit{relaxation function} for computing dissimilarity of concepts defined in $\mathcal{EL}$. However, it can be proved that \textit{triangle inequality} is not always valid and hence, is not a necessary condition. 

\section{Preliminaries} 
\subsection{$\mathcal{ALCH+}$ - A Description Logics Fragment}
We hereby define the semantic interpretation of \lang (an extension of which also interprets current OWL 2 specification)\footnote{The syntax of \lang follows conventional DL as defined in \cite{Baader2003}}. Let $\mathcal{I}$ be an interpretation function, and $\Delta$ be the universal domain. \lang\interpret is defined as:
\begin{itemize}
	\item $\mathcal{AL}$ (\textit{Attributive Language}):
	\begin{itemize}
		\item \textit{Atomic concept}: $A^\mathcal{I} \subseteq \Delta^\mathcal{I}$
		\item \textit{Role}: $r^\mathcal{I} \subseteq \Delta^\mathcal{I} \times \Delta^\mathcal{I}$
		\item \textit{Atomic Negation}: $\neg A$\interpret = $\Delta$\interpret $\setminus A$\interpret 
		\item \textit{Top Concept}: $\top = \Delta$\interpret
		\item \textit{Bottom Concept}: $\bot = \emptyset$\interpret
		\item \textit{Conjunction}: $(C \sqcap D)$\interpret $= C$\interpret $\cap D$\interpret
		\item \textit{Value Restriction}: $(\forall R.C)$\interpret $= \{ a \mid \forall b.(a,b) \in R$\interpret $\rightarrow b \in C$\interpret $\}$
		\item \textit{Limited Existential Restriction}: $(\exists R.\top)$\interpret $= \{ a \mid \exists b.(a,b) \in R$\interpret $\wedge \ b \in \Delta$\interpret $\}$
	\end{itemize}
	\item $\mathcal{E}$ (\textit{Full Existential Restriction}): $(\exists R.C)$\interpret $= \{ a \mid \exists b.(a,b) \in R$\interpret $\wedge \ b \in C$\interpret $\}$
	\item $\mathcal{C}$ (\textit{Concept Negation}): $\neg C$\interpret = $\Delta$\interpret $\setminus C$\interpret
	\item $\mathcal{H}$ (\textit{Role Hierarchy}): $R_1$\interpret $\subseteq R_2$\interpret
	\item $\mathcal{R}_{\mathcal{UNION}}$ (\textit{Role Union}): $R_1$\interpret $\cup \ R_2$\interpret
	\item $\mathcal{R}_{\mathcal{INTERSECTION}}$ (\textit{Role Intersection}): $R_1$\interpret $\cap \ R_2$\interpret
\end{itemize}

\subsection{Formal Similarity Measure}
In this section we define the algebraic properties of $\sigma$ as given in \cite{lehmann2012framework}. Let $C$ is a DL concept in a given terminology (T-Box) $\mathcal{T}$.

\textbf{Definition 1}: A semantic similarity measure $\sigma$ is a function defined as follows:
\\ $\sigma: C \times C \mapsto [0,1]$ where $C \in \mathcal{T}$

\textbf{Properties of Similarity Measure:} Arguably\footnote{* denotes that the property is not universally adopted as necessary condition. Also it cannot be proven to be valid in all types of algebraic spaces.}, $\sigma$ should hold the following properties:
\begin{align}
\textbf{Positiveness*}: \forall C_i,C_j \in \mathcal{T}, \sigma(C_i, C_j) \geq 0
\end{align}
\begin{align}
\textbf{Reflexive}: \forall C_i \in \mathcal{T}, \sigma(C_i, C_i) = 1
\end{align}
\begin{align}
	\textbf{Maximality}: \forall C_i, C_j, C_k \in \mathcal{T}, \sigma(C_i, C_i) \geq \sigma(C_j, C_k)
\end{align}
\begin{align}
	\textbf{Symmetry*}: \forall C_i, C_j \in \mathcal{T}, \sigma(C_i, C_j) = \sigma(C_j, C_i)
\end{align}
\begin{align}
\begin{aligned}
\textbf{Equivalence Closure}: \forall C_i, C_j \in \mathcal{T}, \sigma(C_i, C_j) = 1 \\ \Longleftrightarrow C_i \equiv C_j
\end{aligned}
\end{align}
\begin{align}
\begin{aligned}
\textbf{Equivalence Invariance}: \forall C_i, C_j, C_k \in \mathcal{T}, C_i \equiv C_j \\ \implies \sigma(C_i, C_k) = \sigma(C_j, C_k)
\end{aligned}
\end{align}
\begin{align}
\begin{aligned}
\textbf{Structural Dependency}: \forall C^{i}_{n}, C^{j}_{n}, C_i, C_j \in \mathcal{T}, \backepsilon \\ C^{i}_{n} \equiv \bigsqcap_{k \leq n} \ (C_k \sqcap C_i); \\ 
and ,C^{j}_{n}\equiv \bigsqcap_{k \leq n} \ (C_k \sqcap C_j); \\ then, \lim_{x\to\infty}\sigma(C^i_{n}, C^j_{n}) = 1
\end{aligned}
\end{align}
\begin{equation}
\begin{aligned}
\textbf{Subsumption Preservation}: \forall C_i, C_j, C_k \in \mathcal{T}, \\ C_i \sqsubseteq C_j \sqsubseteq C_k \implies \sigma(C_i, C_j) \geq \sigma(C_i, C_k)
\end{aligned}
\end{equation}
\begin{align}
\begin{aligned}
\textbf{Reverse Subsumption Preservation}: \forall C_i, C_j, C_k \in \mathcal{T}, \\ C_i \sqsubseteq C_j \sqsubseteq C_k \implies \sigma(C_j, C_k) \geq \sigma(C_i, C_k)
\end{aligned}
\end{align}
\begin{align}
\begin{aligned}
\textbf{Strict Monotonicity}: \forall C_i, C_j, C_k \in \mathcal{T}, \backepsilon \\ if, \forall C_l \in \mathcal{T}, C_i \sqsubseteq C_l;  C_j \sqsubseteq C_l \implies C_k \sqsubseteq C_l \\ then, \sigma(C_i, C_k) \geq \sigma(C_i, C_j) \\ and \ if, \exists C_m \in \mathcal{T}, \backepsilon C_i \sqsubseteq C_m; C_k \sqsubseteq C_m \ and\  C_j \nsqsubseteq C_m \\ then, \sigma(C_i, C_k) > \sigma(C_i, C_j)
\end{aligned}
\end{align}

It should be noted that the aforementioned necessary properties may not be sufficient and hence, detailed theoretical analysis has to be done on sufficiency. 

\begin{figure}
	\includegraphics[scale = 0.25]{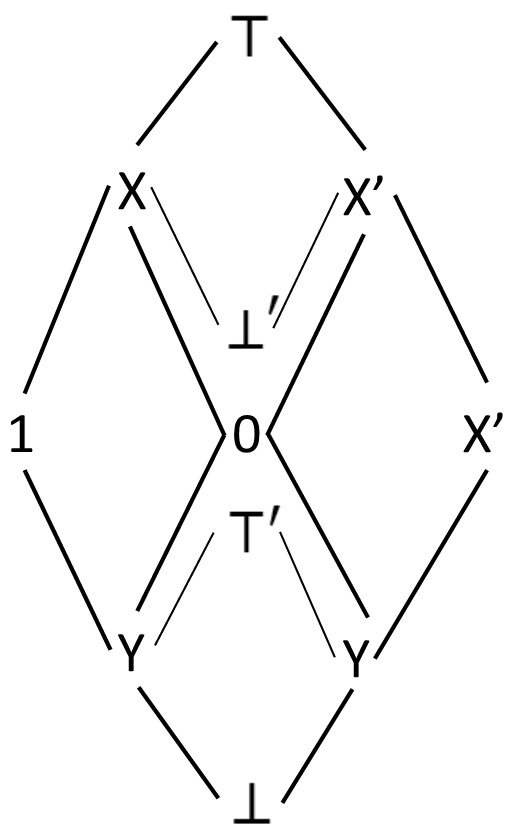}
	\centering
	\caption{Lattice Structure of \alphsim}
\end{figure}

\section{$\mathcal{L_B}$: Formal Language for Concept Coding}
In this section we define the formal language \lsim on which the proposed algebraic interpretation function \isim is defined. We first define \textit{bit} (\alphsim), the alphabet of \lsim, as follows: 

\textbf{Definition 2}: A \textit{base alphabet} ($\Sigma_{\mathcal{B}_{base}}$) is an alphabet defined as: $\Sigma_{\mathcal{B}_{base}} = \{0, 1\}$ where 
\begin{itemize} 
	\item 0 is the empty symbol. It is also called \textit{potential bit} since it generates all other symbols (i.e. bits)
	\item 1 is base bit, called \textit{property bit}, signifying the presence of a property at the string position that it holds.
\end{itemize}

\textbf{Definition 3}: An \textit{bit operator} ($\oplus_\mathcal{B}$) is a set of operators on the base bits defined as: $\oplus_\mathcal{B} = \{\sqcup_\mathcal{B}, \sqcap_\mathcal{B}, \neg_\mathcal{B}\}$.

We now define a very important semantics for \textit{potential bit} (i.e. 0) as follows:
\setcounter{equation}{0}
\begin{align}
\neg_\mathcal{B} 0 = 0
\end{align}

\textbf{Definition 4}: An \textit{derived alphabet} ($\Sigma_{\mathcal{B}_{derived}}$) is an alphabet defined as: $\Sigma_{\mathcal{B}_{derived}} = \{X,X'',X',Y,Y',\top',\bot', \top, \bot\}$ where 
\begin{itemize} 
	\item $X'' = \neg_\mathcal{B} 1$
	\item $X = 1$ \bunion $0$
	\item $X' = X''$ \bunion $0$
	\item $\top' = Y$ \bunion $Y'$
	\item $Y' = 1$ \bintersect $0$
	\item $Y = X''$ \bintersect $0$
	\item $\bot' = X$ \bintersect $X'$
\end{itemize}

Based on the above definition the following observations can be made (using de Morgan's law): 
\begin{itemize} 
	\item $X = \neg_\mathcal{B} Y$
	\item $X' = \neg_\mathcal{B} Y'$
	\item $\top' = \neg_\mathcal{B} \bot'$
	\item $\top = \neg_\mathcal{B} \bot$
\end{itemize}
A further analysis shows that \alphsim has a partial ordering \border \space (as shown in figure 1). It is interesting to note that $b_i$ \bintersect $\neg b_i \neq \bot; \forall b_i \in$ \alphsim.\\
\textbf{Definition 5.1}: An \textit{bit-alphabet} (\alphsim) is defined as \\ \alphsim: $\Sigma_{\mathcal{B}_{base}} \cup \Sigma_{\mathcal{B}_{derived}}$.

It is to be noted that $\oplus_\mathcal{B}$ satisfies commutativity and double negation over \alphsim.

\textbf{Definition 5.2}: A \textit{quantifier-alphabet} ($\Sigma_{\mathcal{B}_\mathcal{Q}}$) is defined as \\ $\Sigma_{\mathcal{B}_\mathcal{Q}} = \{1, 0, X, Y'\}$.
The algebraic space of $\Sigma_{\mathcal{B}_\mathcal{Q}}$ is defined as below:
\begin{itemize} 
	\item $X = 1$ \bunion $0$
	\item $Y' = 1$ \bintersect $0$
	\item $\neg_\mathcal{B} 1 = 0$
	\item $\neg_\mathcal{B} Y' = X$
\end{itemize}
\textbf{Definition 5.3}: A \textit{role-alphabet} ($\Sigma_{\mathcal{B}_\mathcal{R}}$) is defined as \\ $\Sigma_{\mathcal{B}_\mathcal{R}} = \{1, 0, X, Y'\}$.
The algebraic space of $\Sigma_{\mathcal{B}_\mathcal{R}}$ is defined as below:
\begin{itemize} 
	\item $X = 1$ \bunion $0$
	\item $Y' = 1$ \bintersect $0$
\end{itemize}
\textbf{Definition 6.1}: A \textit{base bit-code} (\bcodebase) is defined as \bcodebase $=0^\ast\circ_\mathcal{B}b_i^\ast\circ_\mathcal{B}(\concat||\langle \bconcat b_\forall \concat|| \rangle||\langle \bconcat (b_{\mathcal{R}_i})^+ \concat \rangle|| \bconcat b_i^+)^\ast \bconcat (\concat||\langle \bconcat b_\exists \concat|| (\concat \rangle||\langle \bconcat b_{\mathcal{R}_i})^+ \concat \rangle|| \bconcat b_i^+)^\ast$ where $\circ_\mathcal{B}$ is bit concatenation operator; $b_\forall, b_\exists \in \Sigma_{\mathcal{B}_\mathcal{Q}}; b_\forall = 1, b_\exists = 0$; $b_i \in$ \alphsim; $b_{\mathcal{R}_i} \in \Sigma_{\mathcal{B}_\mathcal{R}}$.\\
\textbf{Definition 6.2}: A \textit{derived bit-code} (\bcodederived) is defined as \bcodederived = (\bcodebase)$^\ast\regbcodederived\bconcat$ (\bcodederived)$^\ast$ where $\concat$ is string concatenation operator.

It can be observed that the definition of \bcodederived is recursive. We leave the explanation and utility of the definition in section 5.3.\\
\textbf{Definition 7}: $\mathcal{L_B}$ is defined as $\mathcal{L_B}=\{$ \bcodebase $\}\cup \{$ \bcodederived$\}$.

\section{Encoding $\mathcal{ALCH+}$ Concept}
\subsection{Motivation}
The motivation behind \textit{BitSim} (\bsim) is to develop a formal, efficient, and scalable matchmaking system that can be applied in DL based knowledge bases. Unlike other DL based similarity measures, \bsim was designed to satisfy all the necessary properties defined in section 3.2 with special emphasis on \textit{structural dependency} and \textit{strict monotonicity}. 

At the same time, \bsim computation is over \lsim, rather than \lang. This significantly improves the computational speed since \bsim essentially becomes a function over \bcode pairs (such bit-codes can be computed and stored off-line in the knowledge base). Since, \bsim computation is performed on pairs of bits holding the same position in \bcode, therefore we can chunk bit-codes in constant sizes and perform similarity over concept pairs on parallel computational platforms. This gives massive scalability to \bsim. Efficient optimization can be performed by caching similarity results of bit-code chunks that are frequently visited.   

We will also show that, at a bit level, \bsim has a partial ordering \bsimorder. This allows application-oriented assignment of similarity scores to bit pairs at the lowest granularity. Also, \bsim is highly adaptive to all types of similarity measures that have set theoretic operations on DL concepts. 

\begin{figure}
	\includegraphics[scale = 0.25]{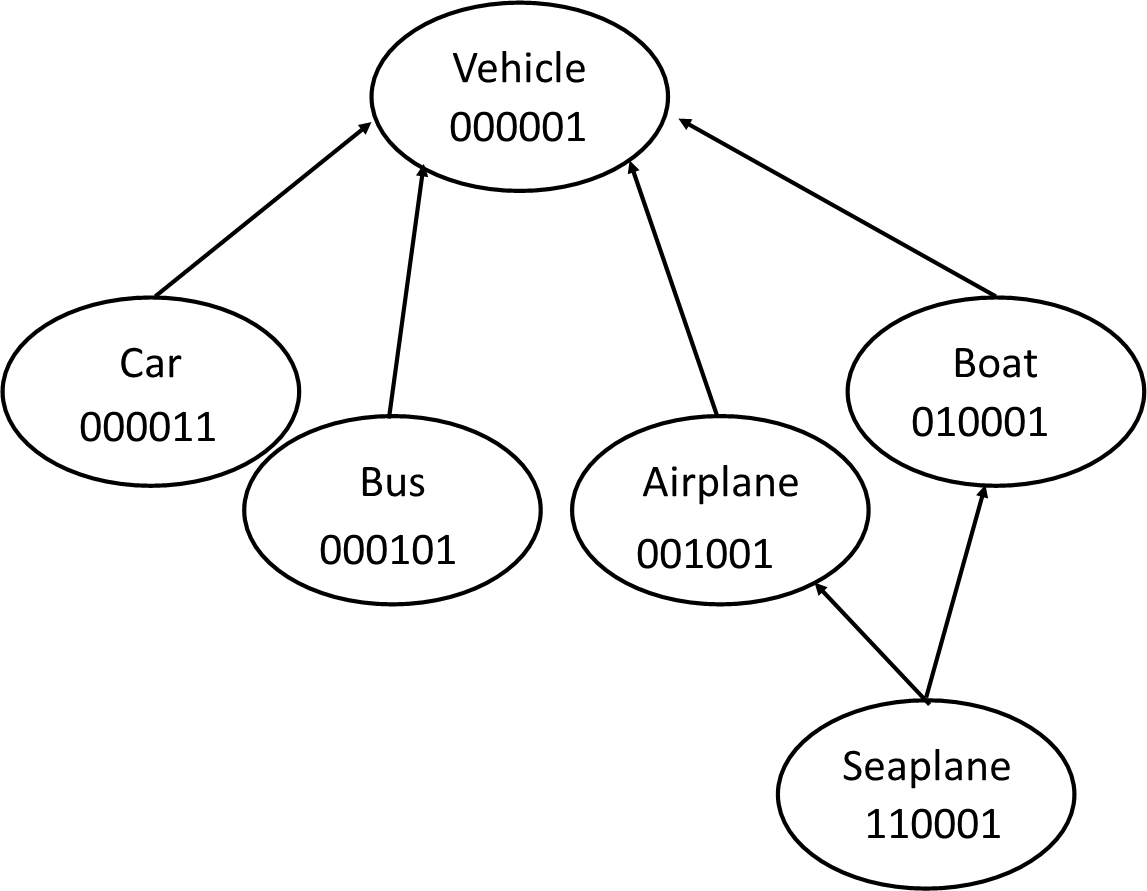}
	\centering
	\caption{Atomic Concept Encoding: Example}
\end{figure}

\subsection{Encoding Atomic Concepts}
Before we show that \lsim has complete correspondence with \lang, we first provide the foundational axioms that helps us to encode atomic concepts in \lang. For that we need to define the proposed algebraic interpretation function \isim (also called \textit{bit-interpretation}). 

\textbf{Definition 8}: \textit{Bit-interpretation} (\isim) is a function as follows:\\
\isim: $C \mapsto$ \lsim; $C \in \mathcal{T_{ALCH+}}$ 

We hereby define $(A_{i})^{\mathcal{I_B}}$, where $A_{i}$ is an arbitrary atomic concept in \lang, using the following two axioms:
\setcounter{equation}{0}
\begin{align}
\begin{aligned}
\textbf{Fundamental Axiom of Atomic Concepts} \\ \forall b_{k}^{A_i} \in \omega^{A_i}_{\mathcal{B}}, \ b_{k}^{A_i} \in \Sigma_{\mathcal{B}_{base}}; \omega^{A_i}_{\mathcal{B}} \equiv (A_{i})^{\mathcal{I_B}}.
\end{aligned}
\end{align}
\begin{align}
\begin{aligned}
\textbf{Axiom of Significant Property Bit} \\ \exists b_{k}^{A_i} \in \omega^{A_i}_{\mathcal{B}}, b_{k}^{A_i} = 1\  \& \ b_{k+1}^{A_i} = 0^\ast.
\end{aligned}
\end{align}
where, \textit{k} is the \textit{k-}th position (in increasing order from right to left) in $\omega^{A_j}_{\mathcal{B}}$. In the second axiom $b_{k}^{A_i}$ holds the significant property bit.

We now show the method to encode inclusion axioms on atomic concepts using the following two axioms:
\begin{align}
\begin{aligned}
\textbf{Axiom of Property Bit Inheritance} \\ \forall A_{i}, \ A_{j} \in \mathcal{T_{ALCH+}}, \backepsilon A_{i} \sqsubseteq A_{j}, \\ b_{k}^{A_j} = 1 \rightarrow b_{k}^{A_i} = 1; \forall b_{k}^{A_j} \in \omega^{A_j}_{\mathcal{B}}
\end{aligned}
\end{align}
\begin{align}
\begin{aligned}
\textbf{Axiom of Atomic Bit-Code Uniqueness} \\ Given, \forall A_{i}, A_{j} \in \mathcal{T_{ALCH+}}, \backepsilon A_{j} \nsqsubseteq A_{i} \\ \exists b_{k}^{A_i} \in \omega^{A_i}_{\mathcal{B}} , b_{k}^{A_j} \in \omega^{A_j}_{\mathcal{B}}, b_{k}^{i} = 1\  \& \  b_{k}^{j} \neq 1	
\end{aligned}
\end{align}

The above four axioms ensure that a simple boolean intersection between any pair of $\omega^{A_i}_{\mathcal{B}}$ and $\omega^{A_j}_{\mathcal{B}}$ generates the bit-code of the \textit{least common subsumer} (\textit{lcs}) atomic concept. An example encoding instance has been illustrated in figure 2. For atomic roles we can use the same principle of encoding with an alphabet $\Sigma^\mathcal{R}_{\mathcal{B}_{base}}$ that corresponds to $\Sigma_{\mathcal{B}_{base}}$. 

\begin{figure}
	\includegraphics[scale = 0.8]{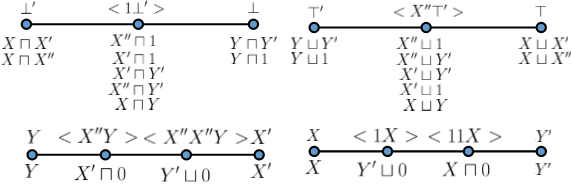}
	\centering
	\caption{Algebraic Space of \compbit}
\end{figure}

\subsection{Encoding Derived Concepts}
For encoding derived concepts, we cannot attain completeness  using \bcodebase only. This is because, for a bounded number of atomic concepts (say, \textit{n}) we need a mechanism to encode $2^{2^n}$ distinct and disjoint concepts, in the worst case. However, with only 11 bits in \alphsim, we can only encode $11^{n}$ distinct concepts. It is because of this reason that we need to use \bcodederived. The method is to have nested encoding for bit operations over certain special bit pairs. These operations do not have direct mapping to the algebraic lattice shown in figure 1. Instead they map to intermediate and discrete \textit{compound bits} (\compbit).  which can be represented in terms of \alphsim. We define \compbit as follows:

\textbf{Definition 9}: A \textit{compound bit} is defined as: \\ \compbit = $\langle\bconcat$\bcodederived$\concat \rangle$, where the algebra of \compbit \space is defined as shown in figure 3.

For any derived concept $C_i$, we can state the following: 
\begin{align}
\begin{aligned}
\textbf{Axiom of Binary Concept Operation} \\ 
b_{\omega_k}^{C_i} \oplus b_{\omega_k}^{C_j} \equiv (\langle \bconcat \ \omega_{\mathcal{B}_{derived}}\concat \rangle)^{C_i}_k \oplus (\langle \bconcat \ \omega_{\mathcal{B}_{derived}} \concat \rangle)^{C_j}_k \\ \equiv (\langle \bconcat \ (\omega_{\mathcal{B}_{derived}}^{C_i} \oplus \ \omega_{\mathcal{B}_{derived}}^{C_j}) \ 	\concat \rangle)_k \equiv b_{\omega_k}^{C_i \ \oplus \ C_j}
\end{aligned}
\end{align}

Based on axiom 5 we can state that:
\begin{lemma}
For any binary operation between \compbit$^{C_i}_k$  and \compbit$^{C_j}_k$, the length of both the operands, \compbit$^{C_i}_k$  and \compbit$^{C_j}_k$, must be same; where \textit{k}: \textit{k-th} position of \compbit \space in \bcode.
\end{lemma}
\begin{lemma}
For any binary operation between \compbit$^{C_i}_k$  and \compbit$^{C_j}_k$, the length of the resultant \compbit$^{C_i \ \oplus \ C_j}_k$ has growth = $O(c^n)$, where $c \in \{1, 2, 3\}$ and \textit{n} is length of operand \compbit.
\end{lemma}
\begin{lemma}
If $\exists b_k \in \omega_\mathcal{B} \backepsilon b_k = \top$, then $\omega_\mathcal{B} = \top$
\end{lemma}
\begin{lemma}
	If $\exists b_k \in \omega_\mathcal{B} \backepsilon b_k = \bot$, then $\omega_\mathcal{B} = \bot$
\end{lemma}
\begin{lemma}
	If $\forall b_k \in \omega_\mathcal{B} \backepsilon b_k = \top'$, then $\omega_\mathcal{B} = \top$
\end{lemma}
\begin{lemma}
	If $\forall b_k \in \omega_\mathcal{B} \backepsilon b_k = \bot'$, then $\omega_\mathcal{B} = \bot$
\end{lemma}
We now postulate the following axioms for operations over derived concepts:
\begin{align}
\begin{aligned}
\textbf{Axiom of Concept Negation}\\
(\neg C_i)^{\mathcal{I}_{\mathcal{B}}} \equiv \{\neg b_{k}^{C_i}\} \equiv \{b_{k}^{\neg C_i}\} 
\end{aligned}
\end{align}
\begin{align}
\begin{aligned}
\textbf{Axiom of Binary Concept Operation} \\ 
(C_i \oplus C_j)^{\mathcal{I}_{\mathcal{B}}} \equiv (C_i)^{\mathcal{I}_{\mathcal{B}}} \oplus (C_j)^{\mathcal{I}_{\mathcal{B}}} \equiv \{b_{k}^{C_i}\} \oplus_\mathcal{B} \{b_{k}^{C_j}\} \\ \equiv  \{b_{k}^{C_i \ \oplus \ C_j}\}; where \ \oplus_\mathcal{B} \equiv \oplus^{\mathcal{I}_{B}}; \oplus \in \{\sqcup, \sqcap\}
\end{aligned}
\end{align}
\begin{align}
\begin{aligned}
\textbf{Axiom of Universal Role Restriction}\\
(\forall R_j . C_i)^{\mathcal{I}_{\mathcal{B}}} \equiv ||\langle \bconcat b_{k}^{\forall} \concat \rangle|| ||\langle \bconcat \{b_{\mathcal{R}_k}^{R_j}\} \concat \rangle ||\bconcat \{b_{k}^{C_i}\}
\end{aligned}
\end{align}
\begin{align}
\begin{aligned}
\textbf{Axiom of Full Existential Role Restriction}\\
(\exists R_j . C_i)^{\mathcal{I}_{\mathcal{B}}} \equiv ||\langle \bconcat b_{k}^{\exists} \concat \rangle|| ||\langle \bconcat \{b_{\mathcal{R}_k}^{R_j}\} \concat \rangle ||\bconcat \{b_{k}^{C_i}\}
\end{aligned}
\end{align}

It is to be noted that for the above axioms $b_k$ can be both a simple and a compound bit. We now provide a proof for showing the mathematical correspondence between (\lsim)$^{\mathcal{I}_{\mathcal{B}}}$ and (\lang)$^\mathcal{I}$.
\begin{lemma}
$\forall b_{k}^{C_i} \in \omega^{C_i}_{\mathcal{B}}, b_{k}^{C_j} \in \omega^{C_j}_{\mathcal{B}};$  \textit{k}: \textit{k}-th position in \bcode \\ $b_{k}^{C_i}$ \border \space $b_{k}^{C_j}$
$\iff (C_i)^{\mathcal{I}_{\mathcal{ALCH+}}} \subseteq (C_j)^{\mathcal{I}_{\mathcal{ALCH+}}}$
\begin{proof}
	Proof can be derived from the lattice structure of \alphsim \space (see figure 1.)
\end{proof}
\end{lemma}
Following the above lemma we can state that:
\begin{theorem}
$(C_i)^{\mathcal{I}} \subseteq (C_j)^{\mathcal{I}} \iff (C_i)^{\mathcal{I}_{\mathcal{B}}} \subseteq (C_i)^{\mathcal{I}_{\mathcal{B}}}$ 
\end{theorem} 
\begin{theorem}
$(C_i \oplus_\mathcal{B} C_j)^{\mathcal{I}} \iff (C_i \oplus_\mathcal{B} C_i)^{\mathcal{I}_{\mathcal{B}}}; \oplus_\mathcal{B} \in \{\sqcup, \sqcap\}$
\begin{proof}
	$\forall b_{k}^{C_i} \in \omega^{C_i}_{\mathcal{B}}, b_{k}^{C_j} \in \omega^{C_j}_{\mathcal{B}}; \\ b_{k}^{C_{i \ \oplus_\mathcal{B} \ j}} = b_{k}^{C_i} \oplus_\mathcal{B} b_{k}^{C_j} \iff b_{k}^{C_{i \ \oplus_\mathcal{B} \ j}}$ \border \space $b_{k}^{C_j}; b_{k}^{C_{i \ \oplus_\mathcal{B} \ j}}$ \border \space $b_{k}^{C_j}$, if $\oplus_\mathcal{B} = \sqcap$; else converse over \border.
\end{proof} 
\end{theorem}
\begin{theorem}
	$(\mathcal{Q} R . C_i)^{\mathcal{I}} \iff (\mathcal{Q} R . C_i)^{\mathcal{I}_{\mathcal{B}}}; \mathcal{Q} \in \{\forall, \exists\}$
	\begin{proof}
	From axiom 7 and 8, we can show that $(\mathcal{Q} R . C_i)^{\mathcal{I}_{\mathcal{B}}}$ is unique. This is because $\mathcal{Q}^{\mathcal{I}_{\mathcal{B}}}$ is unique. In other words, the algebra has complete correspondence with $(\mathcal{Q} R . C_i)^{\mathcal{I}}$.
	\end{proof} 
\end{theorem}
\begin{theorem}
	$\forall$ \bcode $\in$ \lsim, \bcode is unique.
\begin{proof}
Follows from axiom 4 and lemma 1 - 6.
\end{proof}
\end{theorem}
\begin{theorem}
	\lang$^\mathcal{I}$ has complete correspondence with \lsim$^{\mathcal{I}_{\mathcal{B}}}$ 
	\begin{proof}
		The proof follows from theorem 1 - 3. 
	\end{proof} 
\end{theorem}

\begin{figure}
	\includegraphics[scale = 0.8]{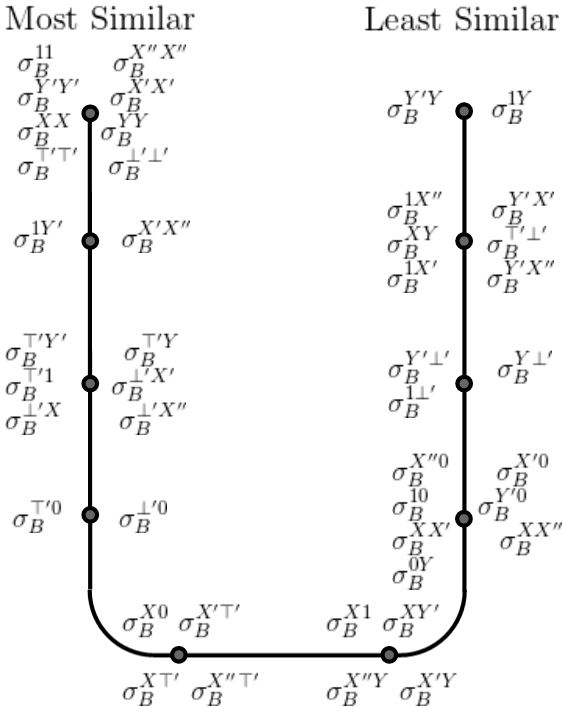}
	\centering
	\caption{Total Ordering of \bsim}
\end{figure}

\section{BitSim Similarity Measure}
\subsection{Outline}
In this section we provide a generic definition for \textit{BitSim}. We first define \bsim (i.e. similarity at a bit level) as follows:

\textbf{Definition 10}: \bsim: $b_i\times b_j \mapsto [0, 1]$; where $b_i, b_j \in$ \alphsim.

It is to be noted that $b_i$ can be \compbit \space as well. One can see that \bsim has a total order \bsimorder (see figure 4). In order to compute similarity at a bit-code level, we define an aggregation function called \bcapsim. There are two parameters that should influence the value output of \bcapsim: (i) \border \space and (ii) \textit{code-generativity} ($\mathcal{CG}$). We  define \textit{code-generativity} as follows:

\textbf{Definition 11}: \textit{Code-generativity} ($\mathcal{F}_{\mathcal{CG}}$) of any \bcode of a concept is the total number of distinct and disjoint concepts that are covered by the \bcode.

As an example, $\mathcal{F}_{\mathcal{CG}}(XX1) = 3$; (i.e. $11'1, 1'11, 111$). We now define the similarity measure at a bit-code level (we call it $\hat{\sigma}_{\mathcal{BS}})$ as follows:

\textbf{Definition 12}: $\hat{\sigma}_{\mathcal{BS}}$: \{\bsim$_k$ $\times \ (\mathcal{F}_{\mathcal{CG}})_k \ \times \ $ \border$_k$\} $\mapsto [0, 1]$.\\

We now postulate the following axioms:
\begin{align}
\begin{aligned}
\textbf{Axiom of 0-bit Similarity} \\
\forall k; \sigma_{BS}(0_k, 0_k) \ is \ ignored.
\end{aligned}
\end{align}
\begin{align}
\begin{aligned}
\textbf{Axiom of} \ \top- \ \textbf{bit Similarity} \\
\forall a \in \Sigma_\mathcal{B}; \sigma_{BS}(\top, a) = 1 \ if \ a = \top; \ else \ undefined.
\end{aligned}
\end{align}
\begin{align}
\begin{aligned}
\textbf{Axiom of} \ \bot- \ \textbf{bit Similarity} \\
\forall a \in \Sigma_\mathcal{B}; \sigma_{BS}(\bot, a) = 1 \ if \ a = \bot; \ else \ undefined.
\end{aligned}
\end{align}

\subsection{$\mathcal{\hat{\sigma}_{BS}}$: Property Analysis}
In this section we show that \bcapsim \space follows the necessary conditions: (i) reflexive, (ii) maximality, (iii) equivalence closure, (iv) equivalence invariance, (v) structural dependency, (vi) subsumption preservation, (vii) reverse subsumption preservation, and (viii) strict monotonicity. The first two properties trivially hold true. The following theorems show that the rest of the properties hold:
\begin{theorem}
Equivalence closure and invariance holds true for \bcapsim.
\begin{proof}
Follows from theorem 1 and theorem 4.
\end{proof}
\end{theorem}
\begin{theorem}
Structural dependency holds true for \bcapsim.
\begin{proof}
Under the condition of structural dependency, for two concepts $C_i$ and $C_j$ the \bcode that is generated for them will have a length, say \textit{l}, that grows exponentially as the number of inner intersections in the definition of both $C_i$ and $C_j$ tends to $\infty$. Therefore, except for the word length of the invariant concepts in the definition of $C_i$ and $C_j$, all of \bcode$^{C_i}$ and \bcode$^{C_j}$ will be exact. Hence, \bcapsim($C_i, C_j$) $\to 1$ 	
\end{proof}
\end{theorem}
\begin{theorem}
Strict monotonicity holds true for \bcapsim.
\begin{proof}
When more than one concepts (say, $C_x, C_y$) subsume two (say, $C_i, C_j$) out of three arbitrary concepts (say, $C_i, C_j, C_k$), while only one (say, $C_y$) subsumes all three, then \bcapsim($C_i, C_j$) $>$ \bcapsim($C_i, C_k$). This is because, since \bcapsim \space is property based measure, $C_i, C_j$ will inherit more common bits than $C_k$. 
\end{proof}
\end{theorem}
We now show how \bcapsim can be adapted to a third-party similarity measure as follows:

\textbf{Definition 13}: $\mathcal{\hat{\sigma}_{BS}}$-$Jaccard =$ \bcapsim($\omega_{BS}^{C_i \sqcap C_j}, \omega_{BS}^{C_i \sqcup C_j}$)

\begin{figure}
	\includegraphics[scale = 0.5]{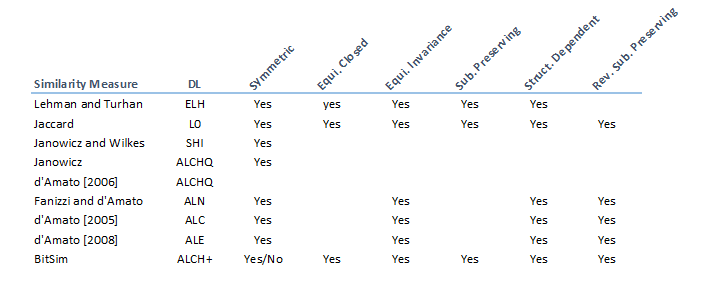}
	\centering
	\caption{Comparative Analysis of BitSim}
\end{figure}

\section{Discussion}
As can be seen, \bsim can be adapted as an alternate paradigm for DL subsumption reasoning. Since \lsim can be mapped to a boolean space one can perform bit operations at high speed and that too on a distributed and parallel platform. At the same time, various caching techniques can be applied efficiently. In the future we will be exploring these research prospects and other possibilities such as probabilistic reasoning on \lsim, A-Box reasoning, and reasoning over higher expressive DL.

\section{Conclusion}
In this paper we have proposed \textit{BitSim} (\bsim) - an algebraic similarity measure for concept definitions in \lang. We show that \bsim satisfies all the necessary algebraic properties recommended for a formal similarity measure. Being based on \isim, \bsim is highly sensitive to standard DL interpretation. Furthermore, \bsim is highly adaptive to any similarity measure that uses set theoretic operations. 

\bibliographystyle{named}
\bibliography{ijcai15}

\end{document}